\documentclass[letterpaper, 10 pt, conference]{ieeeconf}  % Comment this line out if you need a4paper

\IEEEoverridecommandlockouts                              % This command is only needed if
\usepackage[ruled,norelsize]{algorithm2e}
\usepackage{graphicx}
\usepackage{subcaption}
\usepackage{hyperref}
\usepackage{textcomp}

\title{\LARGE \bf
Recognition of Russian traffic signs in winter conditions. Solutions of the ``Ice Vision'' competition winners.
}

\author{Artem L. Pavlov$^{1}$, Azat Davletshin$^{2}$, Alexey Kharlamov$^{3}$, Maksim S. Koriukin$^{4}$, Artem Vasenin$^{2}$, Pavel Solovev$^{3}$, \\Pavel Ostyakov$^{3}$, Pavel A. Karpyshev$^{1}$, George V. Ovchinnikov$^{5}$,
Ivan V. Oseledets$^{6}$, and  Dzmitry Tsetserukou$^{6}$% <-this % stops a space
\thanks{$^{1}$A. L. Pavlov and P. A. Karpyshev are PhD students with the Space Center, Skolkovo Institute of Science and Technology (Skoltech), Moscow, Russia.
        {\tt\small \href{mailto:artem.pavlov@skolkovotech.ru}{artem.pavlov@skolkovotech.ru}}}%
\thanks{$^{2}$A. Davletshin and A. Vasenin are ML Engineers with the NtechLab company.
{\tt\small \href{mailto:a.davletshin@ntechlab.com}{a.davletshin@ntechlab.com}}
}%
\thanks{$^{3}$A. Kharlamov,  P. Solovev and P. Ostyakov are Research Scientists with the Samsung AI Center, Moscow.}%
\thanks{$^{4}$M. S. Koriukin is a PhD student with the Department of System Analysis, Institute of Computer Science and Telecommunications, Reshetnev Siberian State University of Science and Technology, Krasnoyarsk, Russia.}%
\thanks{$^{5}$G. V. Ovchinnikov is a Research Scientist with the Center for Computational and Data-Intensive Science and Engineering (CDISE), Skoltech.}%
\thanks{$^{6}$I. V. Oseledets and D. Tsetserukou are Full and Associate Professors with the CDISE and the Space Center respectively, Skoltech.}%
}

\begin{document}

\maketitle
\thispagestyle{empty}
\pagestyle{empty}

%%%%%%%%%%%%%%%%%%%%%%%%%%%%%%%%%%%%%%%%%%%%%%%%%%%%%%%%%%%%%%%%%%%%%%%%%%%%%%%%
\begin{abstract}

With the advancements of various autonomous car projects aiming to achieve SAE Level 5, real-time detection of traffic signs in real-life scenarios has become a highly relevant problem for the industry. Even though a great progress has been achieved in this field, there is still no clear consensus on what the state-of-the-art in this field is.

Moreover, it is important to develop and test systems in various regions and conditions. This is why the ``Ice Vision'' competition has focused on the detection of Russian traffic signs in winter conditions. The IceVisionSet dataset used for this competition features real-world collection of lossless frame sequences with traffic sign annotations. The sequences were collected in varying conditions, including: different weather, camera exposure, illumination and moving speeds.

In this work we describe the competition and present the solutions of the 3 top teams.

\end{abstract}

%%%%%%%%%%%%%%%%%%%%%%%%%%%%%%%%%%%%%%%%%%%%%%%%%%%%%%%%%%%%%%%%%%%%%%%%%%%%%%%%
\section{Introduction}

The recognition of traffic signs is a multi-category classification problem with unbalanced class frequencies. It is a highly relevant industrial problem, with autonomous driving being the most prominent application of the technology. While being simpler than many other computer vision problems due to the availability of reference images, it has significant reliability requirements, since false detections may cause incorrect behavior of an autonomous car, which may end in a traffic accident.

Additionally, while traffic signs show a wide range of variations between
classes in terms of color, shape, and the presence of pictograms
or text, some classes contain important information which must be recognized for sign detection to be useful, e.g. speed limit or distance information. Other signs change their meaning depending on orientation, e.g. pedestrian crossing or turn left/right signs. In other words, subtle differences in signs can have a significant impact on decision making.

Moreover, algorithms have to cope with large variations in visual appearance due to illumination changes, partial occlusions, rotations, weather
conditions, scaling, etc. And they have to deal with a large number of highly unbalanced classes, e.g. in Russia there are almost 300 sign classes, not counting variations inside class (i.e. speed limit is counted as a single class). Plus the number of classes rises fast if solution is intended to be used in many countries, most of which have different traffic signs.

According to the survey by Paclik \cite{Paclik}, the works on traffic sign detection and recognition started as early as in 1984. A great deal of excellent algorithms for traffic sign detection was proposed since then. Detailed reviews on this topic can be found in \cite{zhu2017overview} and \cite{janai2017computer}.

\begin{figure}
    \centering
    \begin{subfigure}[b]{0.49\columnwidth}
        \includegraphics[width=\textwidth]{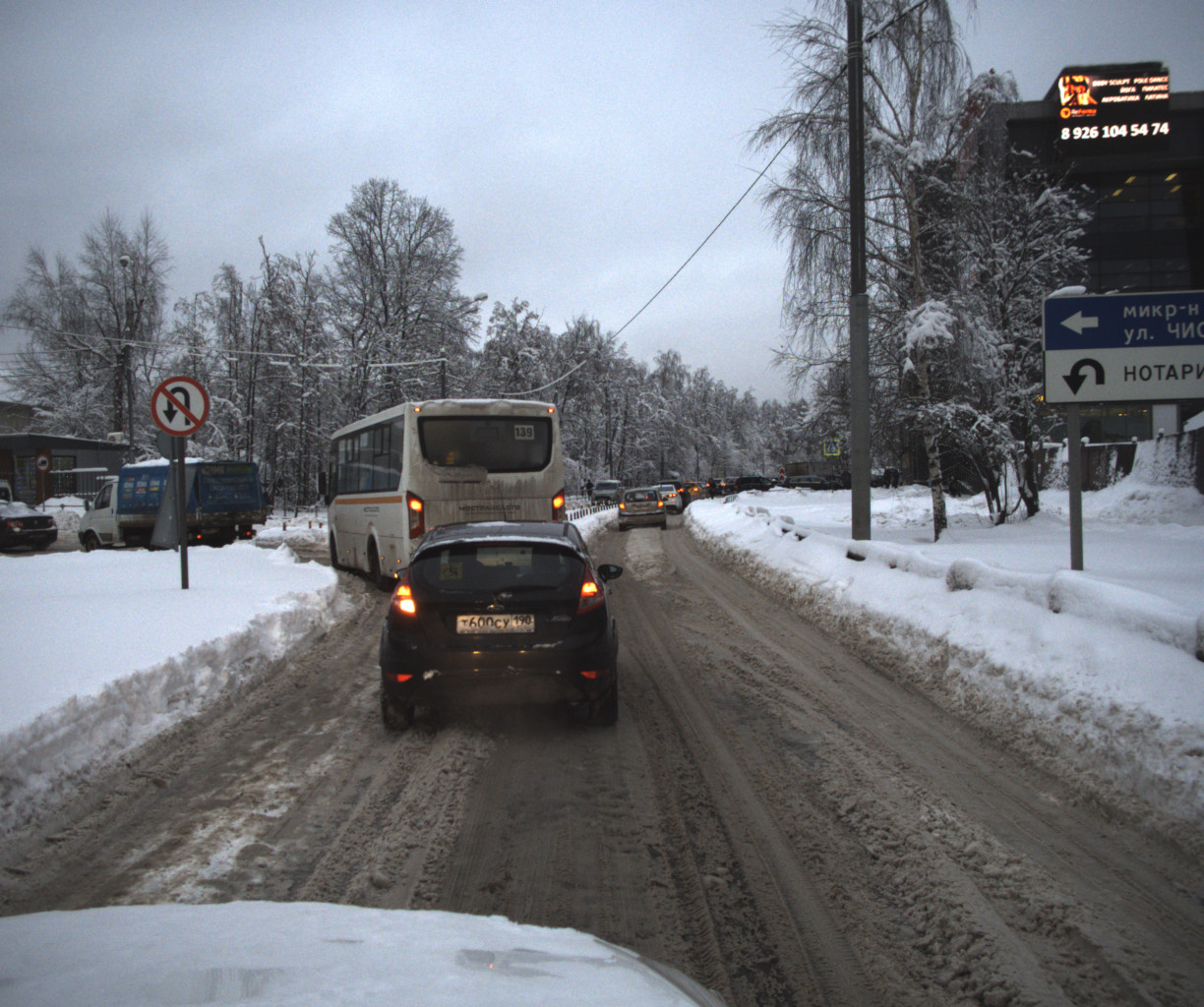}
    \end{subfigure}
    \begin{subfigure}[b]{0.49\columnwidth}
        \includegraphics[width=\textwidth]{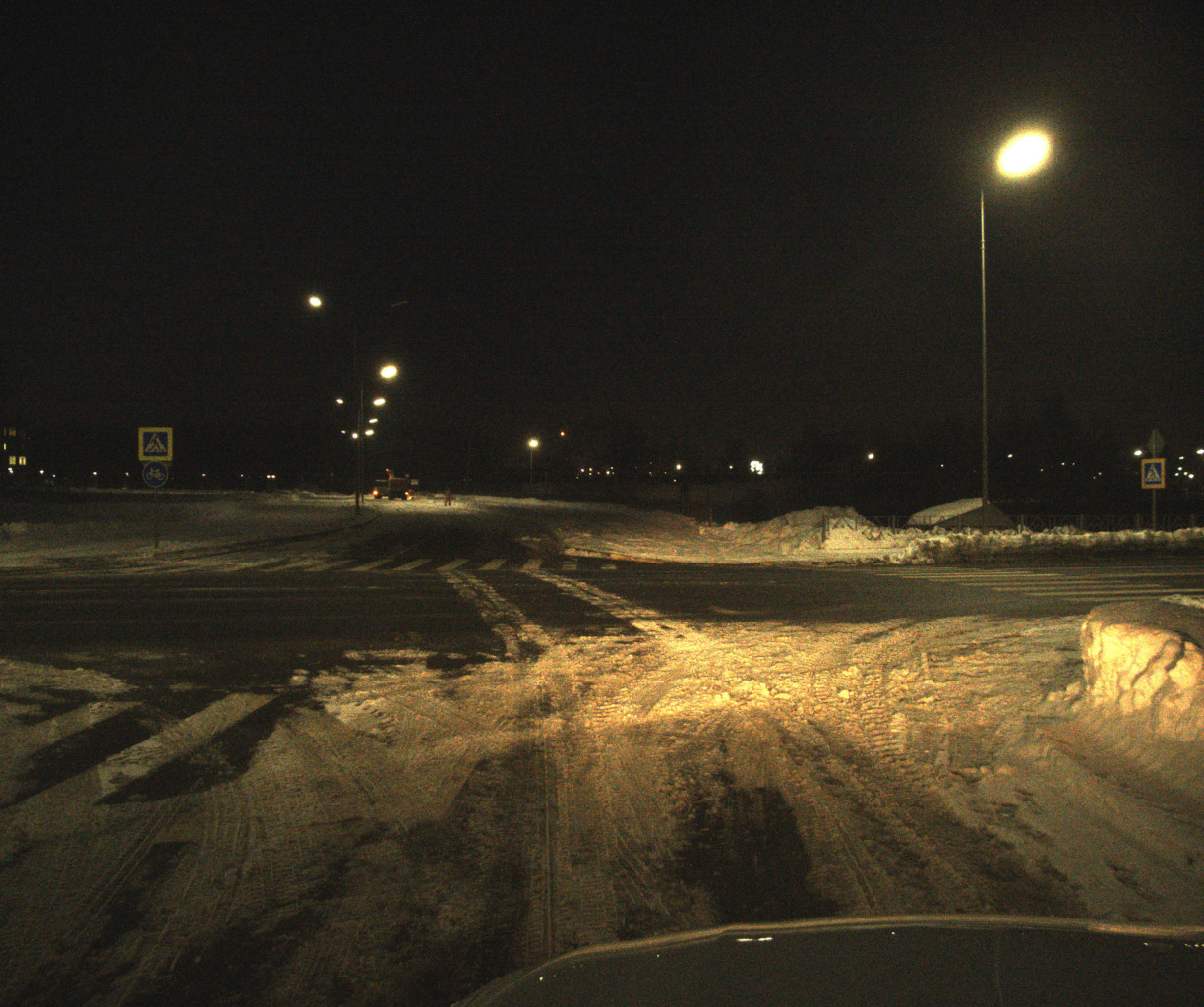}
    \end{subfigure}
    \caption{Frame examples from the IceVisionSet dataset.}
\end{figure}

\begin{figure}
    \includegraphics[height=0.13\textwidth]{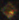}
    \includegraphics[height=0.13\textwidth]{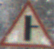}
    \includegraphics[height=0.13\textwidth]{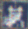}
    \includegraphics[height=0.13\textwidth]{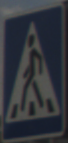}
    \\\\
    \includegraphics[height=0.097\textwidth]{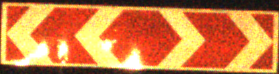}
    \includegraphics[height=0.097\textwidth]{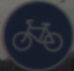}
    \caption{Annotated sign examples from the IceVisionSet dataset.}
\end{figure}

Deep learning approaches are overwhelmingly popular amongst the recent works in this field. Having large datasets is of the utmost importance for training such solutions. A great number of traffic sign datasets for different countries has been previously published, including: the Belgian Traffic Sign Dataset \cite{timofte2014multi}, German Traffic Sign Recognition Benchmark \cite{Stallkamp2012}, German Traffic Sign Detection Benchmark \cite{Houben-IJCNN-2013}, Swedish traffic sign dataset  \cite{larsson2011using}, Chinese Traffic Sign Database\footnote{\url{http://www.nlpr.ia.ac.cn/pal/trafficdata/index.html}} and Russian Traffic Sign Dataset (RTSD) \cite{RTSD}.

One of the most recent traffic signs datasets is IceVisionSet \cite{pavlov2019icevisionset}. It focuses on Russian winter roads and covers both day and night conditions. Contrary to many datasets, it provides sequences of raw Color Filter Array (CFA) frames compressed using lossless algorithms. This dataset was used for the ``Ice Vision'' competition, the final stage of which was held in July 2019.

In this paper we present a short description of the competition and the solutions of the 3 top teams:
\begin{itemize}
\item First place: Azat and Artem (Azat Davletshin, Artem Vasenin).
\item Second place: PsinaDriveNow (Alexey Harlamov, Pavel Solovev, Pavel Ostyakov).
\item Third place: Vizorlabs (Maksim S. Koriukin).
\end{itemize}

\section{The ``Ice Vision'' competition}

\begin{figure*}
	\centering
    \includegraphics[width=0.97\textwidth]{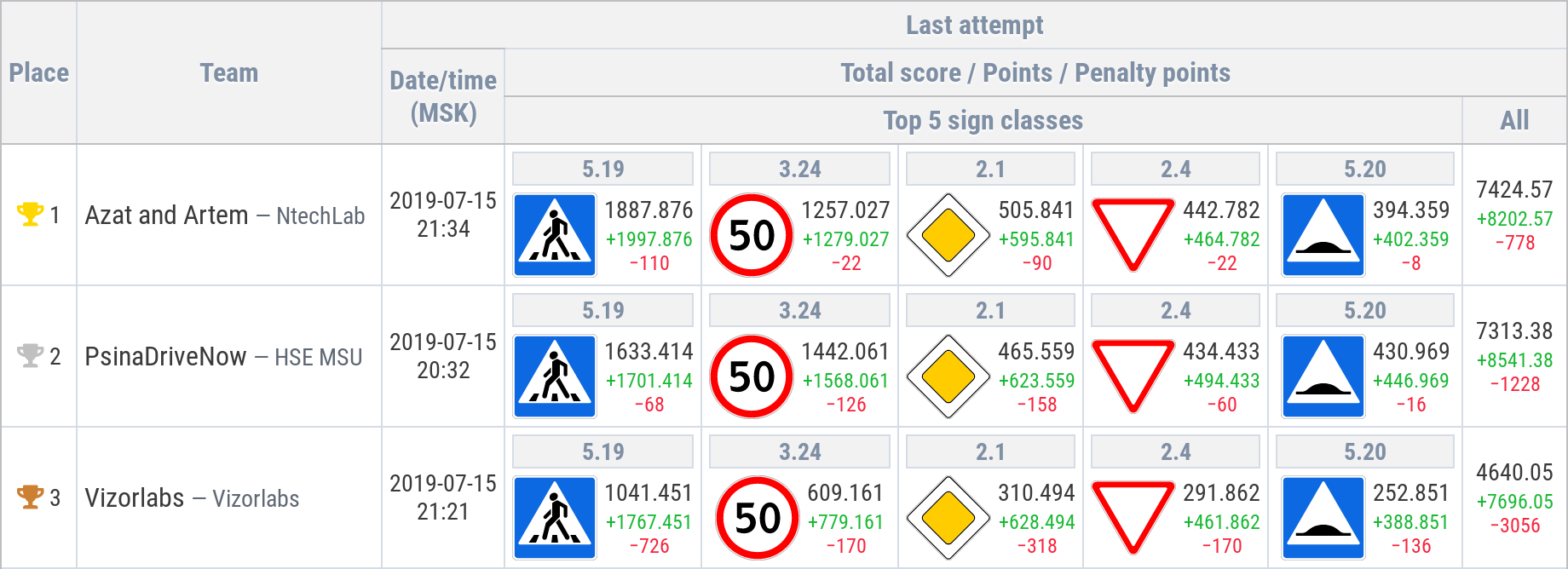}
    \caption{Part of the ``Ice Vision'' leaders table available at \href{https://visiontest.misis.ru}{https://visiontest.misis.ru}. Green numbers denote score for successful detections and red numbers denote penalties for incorrect detections.}
\end{figure*}

The competition was organized by the Russian Venture Company in association with the National University of Science and Technology MISiS, Skolkovo Institute of Science and Technology , and the ``Starline'' company. It was divided into 2 stages: online and offline. The first stage was used to select teams which will participate in the second one. Additionally, several foreign teams have been invited to the offline stage, namely: the Harbin Institute of Technology, the Politechnical University of Catalonia, the University of Paris-Saclay, the University of Science and Technology Beijing, the University of Wisconsin–Stout, and the University of Ulsan.

The competition was based on the IceVisionSet dataset. The annotations had two distinct parts. The first part contained frame-by-frame annotations with linearly interpolated bounding boxes created using Computer Vision Annotation Tool (CVAT)\footnote{\url{https://github.com/opencv/cvat}}, this part was presented in \cite{pavlov2019icevisionset}. The second part was annotated after the paper submission. It was done in the \href{supervise.ly}{Supervisely} web-tool, and had only approximately each 30th frame annotated (we have used a random step between 25 and 35 frames, so participants will not know which frames are annotated), but all annotations were manual.

Both dataset images and annotations are published under open CC BY 4.0 license and can be downloaded from \url{http://oscar.skoltech.ru} and \url{https://github.com/icevision/annotations} respectively. You can find annotation statistics behind the second link.

For the competition an IoU-based metric was used. Bounding boxes with an area smaller than 100 pixels were ignored during evaluation. The final result was a sum of points for each correctly detected sign minus penalties for false-positive detections (2 points for each such detection). If a sign was detected twice, then the detection with the smallest IoU was counted as a false positive. Scoring was handled using a scoring software with an open source code, see: \url{https://github.com/icevision/score}.

During the online stage the participants had to detect only 10 sign classes. See table in the README v0.1.5\footnote{\url{https://github.com/icevision/score/tree/v0.1.5}}. Detection was considered successful if IoU was bigger or equal to 0.5 and the bounding box had a correct class code. If IoU of a true positive detection was bigger than 0.85, it resulted in adding 1 point to final result. Otherwise the points were calculated as $((IoU - 0.5)/0.35)^{0.25}$.

During the offline stage the participants had to detect all signs defined by the Russian traffic code. Detection was considered successful if IoU was bigger or equal to 0.3 and the bounding box had a correct class or superclass code. Score for true positives was calculated as $(1 + k_1 + k_2 + k_3)*s$, where: $s$ is the ``base'' score, $k_1$ is a coefficient for detecting sign code, $k_2$ is a coefficient for detecting associated data, $k_3$ is a coefficient for detecting temporary sign. If $(1 + k_1 + k_2 + k_3) < 0$, the detection score was set to 0. If $IoU > 0.85$, $s = 1$. Otherwise it is calculated using the following equation: $((IoU – 0.3)/0.55)^{0.25}$. For a complete description of coefficients computation please refer to the README.

Additionally, the participants were constrained by computational power and time. During the offline stage they had to process ~100 000 frames (~50 000 stereo-pairs) in 5 hours using a virtual machine with a single NVIDIA Tesla V100 and 8 core CPU, provided by Zhores Cluster\footnote{\url{https://www.zhores.net/}}.

All participant's submissions from the online and offline stages are published under open CC BY 4.0 license and can be downloaded from: \url{https://github.com/icevision/solutions}.

\section{The First Place Solution}

\subsection{Algorithm overview}
The team used a ``detect and track'' approach for their final solution.
The detector architecture was Faster R-CNN\cite{ren2015faster} with FPN\cite{FPN}, Cascade Head \cite{Cascade} and Deformable Convolutions\cite{Deformable}.
IoU-Tracker\cite{IoU} was used for tracking.
Each track was further refined by combining class probabilities from all bounding boxes belonging to a track and selecting the best one.
For each track that was classified as a speed limit sign, the team also predicted the speed limit value.

The train set of the competition was relatively small, therefore the team pre-trained the detector on COCO\cite{COCO} and RTSD\cite{RTSD}.

The code of the final solution can be found at: \href{https://gitlab.com/rizhiy/ice_vision}{https://gitlab.com/rizhiy/ice\_vision}.

\subsection{Detection}
The team used MMdetection\cite{chen2019mmdetection} as the base framework for their detection pipeline.
The baseline configuration was Faster R-CNN + FPN, pre-trained on ImageNet\cite{ImageNet}.
Since the challenge had performance and time restrictions, the team decided
 to use \mbox{ResNet-50\cite{ResNet}} as a backbone to balance speed and accuracy.

During the challenge, the team has performed several experiments to see which approaches perform well on the competition dataset.  For experiments evaluation the standard COCO mAP metric was used.

As a starting point the team has used a model pre-trained on the COCO dataset provided by the MMdetection framework. Next model was trained on RTSD, this dataset also contains Russian traffic sign annotations for videos captured from a moving car and is bigger than the training dataset used in the competition. Finally the resulting model was fine-tuned on the training data provided by organizers. The final training pipeline was as follows: COCO $\rightarrow$ RTSD $\rightarrow$ IceVision.

The metric used in this competition awarded higher score to more precise detections (higher IoU),
therefore the team made an experiment using Cascade Head.
Cascade Head uses multiple stages to refine proposals produced by the FPN, each trained with a higher IoU threshold.
This approach produces more precise detections.

While traffic signs have pretty defined shape, when captured from a car they can appear squashed and twisted.
To ease the amount of work the classifier has to do, the team used Deformable convolutions which can bring features into a unified shape.

Multi-scale training augmentation was used, as it almost always helps with lack of data.

Many of the signs in the dataset were small (with an area less than 300 pixels).
JPEG compression negatively impacts classification on small signs.
By using original raw Bayer PNM files provided by the organizers and using bi-linear demosaicing to calculate RGB image expected by the backbone, the team was able to improve classification on small signs.

The final configuration achieved COCO mmAP score of 0.412 on the validation set.

\subsection{Tracking}
Running detector on each frame at good resolution was not feasible because of the computation power and timing constraints.
Therefore the team has decided to run the detector once every three frames, track the detections and linearly interpolate the bounding boxes in-between.
To track the bounding boxes the team used IoU tracker.

IoU tracker works as follows:
For each track, IoU is calculated for all detections in a new frame.
If the highest IoU is higher than the threshold, that detection is added to the track.
All detections which are not assigned to a track, start as a new one.

During testing the team found that signs can move quite a lot even in three frames, but are located sparsely.
Therefore, lower IoU thresholds produced better tracks.
In final solution the team used IoU threshold of 0.1.

\subsection{Track Refinement}
Usage of a tracker also allowed to improve the quality of detections.
In the video the same signs can be seen multiple times at various distances.
By tracking it, the team was able to use predictions from multiple frames.
Predictions were averaged and the best one was assigned to all detections in the track.

Additionally, since competition allowed predicting super-categories, the team added simple logic to combine class probabilities.
The logic was as follows: if none of the most specific class probabilities were above the threshold, class probabilities of each $2^{nd}$-level-category were summed up.
Each probability was again checked if it was above the threshold and if not, probabilities for top-level-categories would be summed up and again checked against the threshold.

The team used different thresholds for each category level.
Thresholds themselves were found using grid-search on validation set.

Finally, if a sign was classified as a speed limit sign, the bounding box was cut out from the original image and the number on the sign was predicted using a ResNet-34.

Since the metric used in the competition summed, rather than averaged, scores of each class, therefore the team did not attempt to fix class imbalance.

\section{The Second Place Solution}
The team used Cascade R-CNN \cite{Cai_2018_CVPR} with the ResNeXt101 \cite{xie2017aggregated} backbone as a base architecture.
Training was done using the MMDetection framework \cite{chen2019mmdetection} in 2 stages:
\begin{enumerate}
    \item Training of the full model (head and backbone) using RTSD dataset \cite{RTSD}.
    \item Fine-tuning using IceVisionSet images after demosaicing, cropping of the lower 600 rows and resizing to 1632x966 pixels. 
\end{enumerate}

Due to the timing restrictions, ResNext101 features were computed only for each 5\textsuperscript{th} frame. Bounding box positions for in-between frames were interpolated using cross-correlation metric.

Source code of the final solution can be downloaded from \href{https://github.com/gamers5a/SignDetection}{https://github.com/gamers5a/SignDetection}.

\subsection{Interpolation}

\begin{figure*}
    \includegraphics[width=0.99\textwidth]{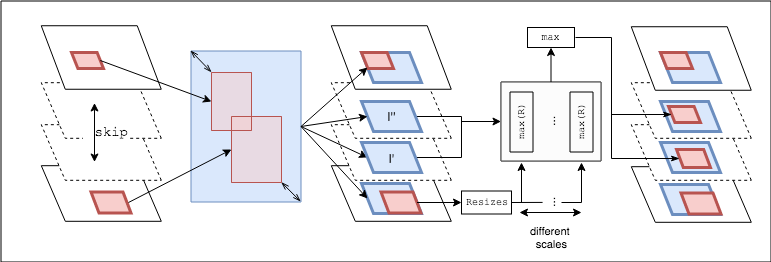}
    \caption{Interpolation pipeline used by the PsinaDriveNow team.}
    \label{fig:psina_tracking}
\end{figure*}

Matching of bounding boxes between keyframes was done using normalized cross-correlation between sub-images selected by bounding boxes, i.e. without taking classification results into account. After a pair of bounding boxes have been found, a search area is found by adding/subtracting 20 pixels from maximum/minimum coordinates of bounding boxes (see Fig. \ref{fig:psina_tracking}).

The size of an in-between bounding box was calculated using linear interpolation, while position was found using maximum cross-correlation response in the search area.

\subsection{Signs classification}

One of the tasks of the competition was the recognition of text on signs (city signs, numerical values of the applicability of the signs.
Since solving the OCR problem by tailoring the conditions of entries provided by the organizers requires a considerable amount of time, the team has decided to approach the solution of the problem from the other side, namely, 
to solve the problem of classifying text values rather than recognition.

One of the features was the inability to directly use flips during training, since there are signs of pedestrian crossings in different directions in the training, the rotation of which changed the annotation to the wrong one, which is why during the training 
the team first used flips, and in the very last era it was turned off for network to learn correct orientation of a sign.

\subsection{Inference speed optimization}

To accelerate the inference, the team used Float16, which is supported by the MMDetection framework, while training was done using Float32. 
This approach allowed to improve the network performance from 2 to 3 fps on the original pictures. During dataset exploration, the team has noted that the camera is installed on the machine in a such way that signs almost never appear in the lower part of frames, which is expected since signs are installed to improve visibility for drivers. Thus the  team has ignored lower 600 out of 2048 rows, it improved detector performance by additional $30\%$.

Since the final testing assumed a large load on the file system implemented using IBM GPFS, the team has decided to reduce the use of random reading from a remote file storage by asynchronously caching video sequences to RAM, in parallel with their prediction, while the GPU and disk load were further balanced using the Round-Robin algorithm. Since the the team solution required 3 independent passes through video sequences (a purely sequential reading of which would take more time than was allocated for predicting the entire test suite), this approach allowed the team to get rid of data loading problems.

\subsection{Domain adaptation}
The dataset contains both day and night sequences. Night sequences were under-exposed in some parts to limit blur caused by the car movement. This is why the team has used histogram equalization filter incorporated into the frame conversion tool provided by the organizers. The histogram equalization  filter boost image contrast, but amplifies noise, degrades colors and changes background brightness depending on the amount of bright objects in the frame.

This is why predictions on night data are less accurate than day data. To improve the results on night sequences, the team has tried to increase dataset diversity by applying Domain Adaptation of day sequences to night ones.
To do this, the team has trained MUNIT \cite{huang2018munit},
to translate day frames to night ones and vice versa. The reverse transformation turned out to be of rather poor quality, because night images simply do not have all the necessary information to restore the image, so only day to night
conversions were added to the training dataset.

\begin{figure*}
    \includegraphics[width=0.99\textwidth]{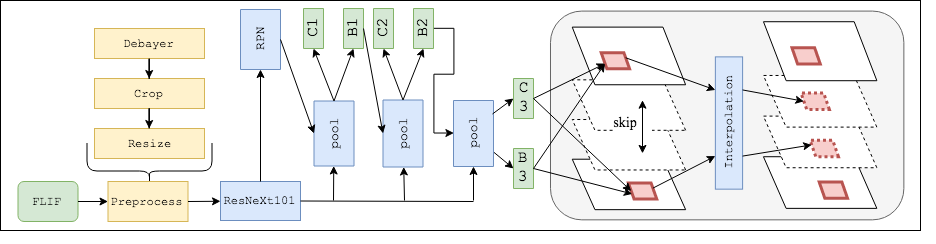}
    \caption{Detection pipeline used by the PsinaDriveNow team. ``B'' stands for ``bounding box'', ``C'' for classification result and RPN for Region Proposal Network.}
    \label{fig:psina_cascade}
\end{figure*}

\section{The Third Place Solution}
Most of the existing methods for object detection and classification can be roughly
divided into three main stages: proposal of a candidate object from the region,
selection of signs and classification. In order to establish the regions that are likely to contain objects of interest, many methods use the sliding window search strategy\cite{girshick2011object}. These methods use windows with different scaling
and ratio to scan the image with a fixed pitch size. An algorithm called ``Selective search'' was proposed by Uuijlings \textit{et al.}\cite{uijlings2013selective} to establish possible locations of target objects. During the selective search, one starts with each individual pixel as its own group. Then, one calculates the texture for each group and combines the two textures that are the closest. The algorithm continues to merge the regions until all groups are clustered.

These methods are widely used with deep convolutional networks for object detection. In \cite{girshick2015fast}, the region proposal network (RPN) was offered, as well as the new neural network architecture for objection detection named R-CNN. The region proposal method (RPN) subsequently became the most popular method for proposing the region for the object of interest. The R-CNN architecture requires many region proposals to be precise, but many regions overlap with each other. Instead of extracting the functions for each image patch from scratch, the functions extractor based on the convolutional neural network can be used \cite{ren2015faster}. The team has  also used an external region proposal method (selective search), to establish the areas of interest that are subsequently combined with relevant object maps to obtain corrections for object detection.

The object detection methods which follow this approach are well known as two-stage methods: proposal of the region candidate at the first stage and object classification at the second one. The two-step methods based on convolutional networks achieve high accuracy on object detection tasks, but have a very low image processing speed. Methods which do not require additional operations region proposals, such as YOLO \cite{redmon2016you} and SSD \cite{liu2016ssd}, represent so-called one-step methods. They are much faster than the two-step methods, but have lower accuracy. In particular, this trade-off makes them unsuitable for detection of large number of small objects, which were common in the competition dataset.

\subsection{Implementation}
\begin{figure}
    \includegraphics[width=\columnwidth]{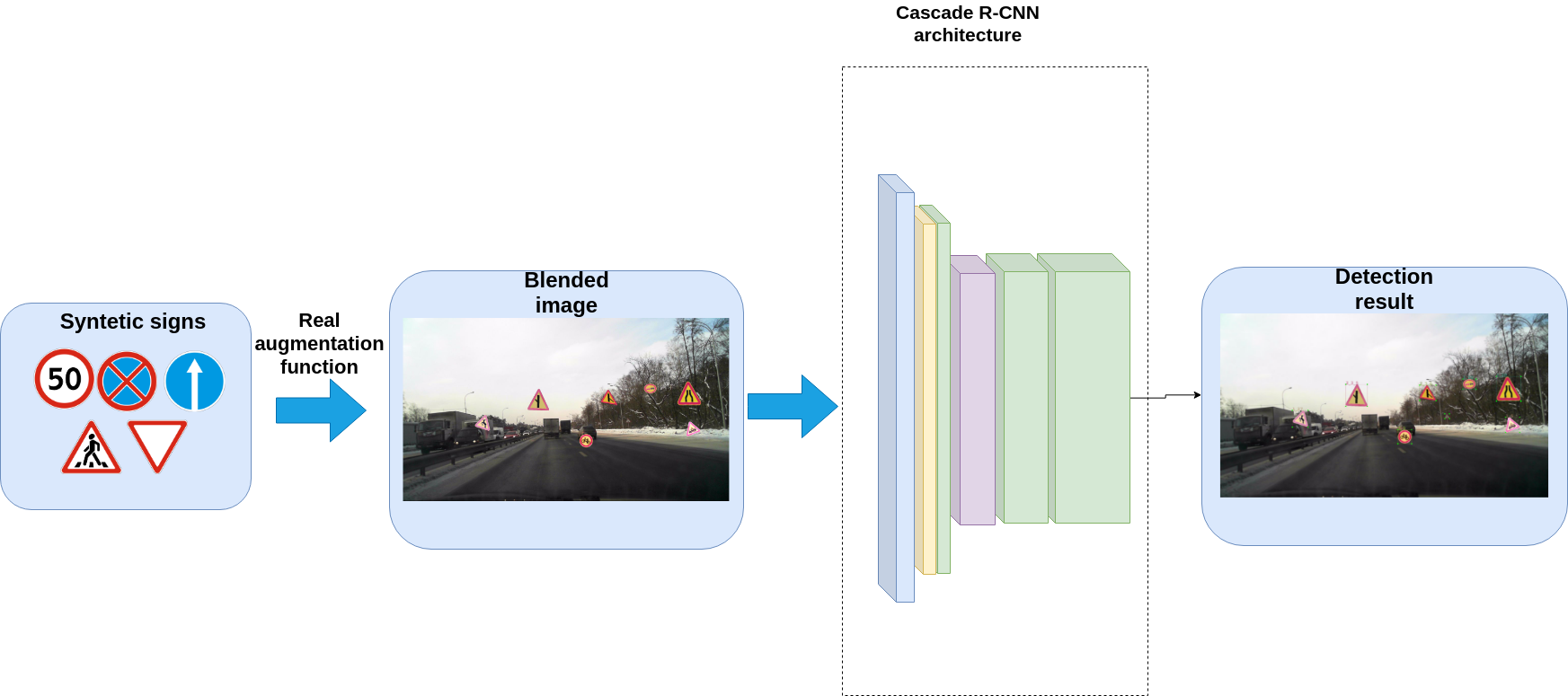}
    \caption{Pipeline used by the VizorLabs team.}
    \label{fig:vizorlabs_diagram}
\end{figure}
Large datasets is a key factor for training well-functioning CNN-based architectures that have millions of parameters. In the past, some well-known datasets for various tasks have been published, for example, ImageNet \cite{deng2009imagenet}.

Due to the small amount of available training data, it was decided to enhance it by augmenting it with synthetic data. The figure below shows the general arrangement for the solution applied. 

\begin{figure}
    \centering
    \begin{subfigure}[b]{0.45\columnwidth}
        \includegraphics[width=\textwidth]{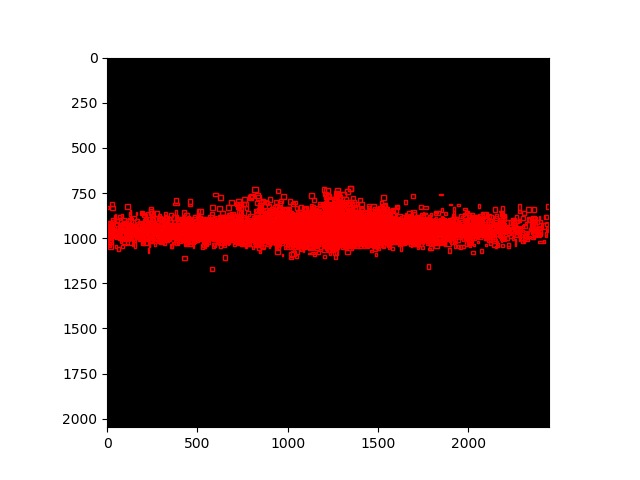}
        \caption{Smaller than 30x30}
    \end{subfigure}
    \begin{subfigure}[b]{0.45\columnwidth}
        \includegraphics[width=\textwidth]{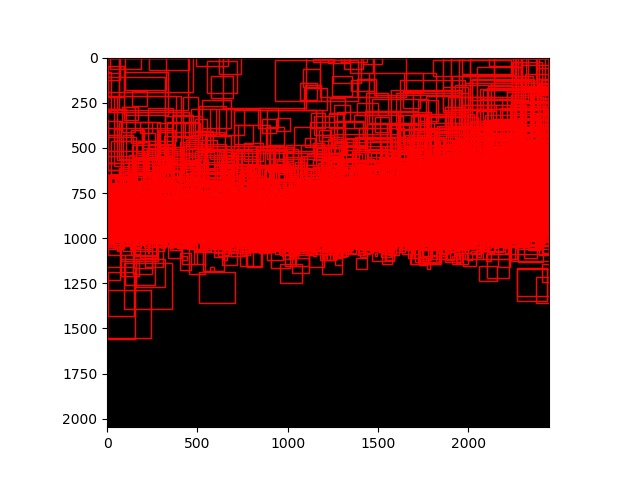}
        \caption{Bigger than 30x30}
    \end{subfigure}
    \caption{Spatial distribution of traffic signs in the IceVisionSet dataset for different sign sizes.}
    \label{fig:vizorlabs_dist}
\end{figure}

The team used Cascade R-CNN as a neural network architecture with usage of the transfer learning method \cite{pan2009survey} using the pre-trained ResNetXt-101 \cite{xie2017aggregated} on the COCO dataset \cite{COCO}, and then set up the model for the IceVision dataset. ResNetXt-101 CNN was used as a backbone. Adam \cite{Kingma2014AdamAM} was used as an optimizer of the objective error function, with a training rate of 5e-3. While preparing the data for training, the team have used the Random Image Cropping approach. Each batch of images included one fragment with the dimensions of 2448 × 1700. The decision to use only part of the image for training was made upon viewing the spatial distribution of traffic signs on all images (see Fig. \ref{fig:vizorlabs_dist}). Objects in a lower half of the frames were extremely rare.

\section{Conclusions}
In this work we have presented the ``Ice Vision'' competition focused on real-time detection of Russian traffic signs in winter conditions. The IceVisionSet used for the competition features lossless real world data collected on Russian public roads in different conditions, including weather and illumination. During the competition strong timing restrictions were enforced upon participants.

This work covers 3 solutions of the competition winners. Two teams have published the source code of their solutions.

We can observe that solutions share some similarities, common for modern deep learning architectures, i.e. solutions use backbone/head architecture and heavily rely on transfer learning, detectors work with single images without utilizing temporal information outside of specialized algorithm-based trackers. Probably due to the reliance on transfer learning solutions have not used raw Bayer images and stereo information.

%\section*{Acknowledgments}
%TODO

\bibliographystyle{IEEEtran}
\bibliography{IEEEabrv,refs0,refs1,refs2,refs3}
\end{document}